\documentclass[10pt,twocolumn,letterpaper]{article}

\usepackage{iccv}
\usepackage{times}
\usepackage{epsfig}
\usepackage{graphicx}
\usepackage{amsmath}
\usepackage{amssymb}
\usepackage{amsfonts}
\usepackage{subfigure}
\usepackage{algorithm}
\usepackage{algorithmicx}
\usepackage{algpseudocode}
\usepackage{booktabs}
\usepackage{multirow}
\usepackage{url}
\usepackage{epstopdf}

\floatname{algorithm}{Procedure}

\DeclareMathOperator*{\argmax}{arg\,max}

\iccvfinalcopy 

\def\iccvPaperID{2651} 

\ificcvfinal\fi
\begin{document}

\title{Sliding Window Optimization on an Ambiguity-Clearness Graph for Multi-object Tracking}

\author{Qi Guo\hspace{3em}Le Dan\hspace{3em}Dong Yin\hspace{3em}Xiangyang Ji\\
Tsinghua University, Beijing, 100084, China
}

\maketitle

%
\begin{abstract}
Multi-object tracking remains challenging due to frequent occurrence of occlusions and outliers. In order to handle this problem, we propose an Approximation-Shrink Scheme for sequential optimization. This scheme is realized by introducing an Ambiguity-Clearness Graph to avoid conflicts and maintain sequence independent, as well as a sliding window optimization framework to constrain the size of state space and guarantee convergence. Based on this window-wise framework, the states of targets are clustered in a self-organizing manner. Moreover, we show that the traditional online and batch tracking methods can be embraced by the window-wise framework. Experiments indicate that with only a small window, the optimization performance can be much better than online methods and approach to batch methods.
\end{abstract}

\section{Introduction}
\label{sec:intro}
With the development of computer vision techniques, more and more people began to focus on understanding the behavior as well as other context of the objects via visual information. Tracking targets in video sequences, one of the core topics with wide applications in video surveillance, rocketed with the boost of tracking-by-detection (TBD) methods \cite{smeulders2014visual}. The TBD reconstruct the states of targets based on the detection responses by assigning identity to each detection and optimizing the trajectories \cite{Milan2012,Milan2014}. The prosperity of TBD these years has raised people's interests in a more challenging topic - multi-object tracking (MOT) with unknown numbers.  MOT remains difficult due to complex settings of sequences, \eg, intricate trajectories of targets, varying illumination, movements of cameras, \etc.

The MOT problem can be handled in an online fashion, which could be adopted in time critical applications. However, the traditional online methods is susceptible to outliers brought by occlusions and noises, \eg, false positives, true negatives, duplicate detections of a single target, \etc. These outliers can cause ambiguities in data association. Some tackles the problem using sparse appearance model \cite{sparserepresentation2013,LSSsparse2013}, and others via prediction \cite{Bae2014} of states in future frames. But dynamics and appearances of the targets are unpredictable in some cases. Batch tracking methods are easier to solve the problem of outliers than online methods by global optimization of association and trajectories. Terms that penalize mutual exclusions and the number of tracklets \cite{Milan2014,choi2015near} were added to the energy function to regularize trajectories.

Apart from advantages of batch methods, one major problem is that the global optimization involves frames in the whole sequence \cite{Bae2014TIP} which does not suit for real-time applications. Some batch methods require initial solutions, \eg \cite{Milan2014}. Therefore, we propose our method in this paper, aiming at combining advantages of online and batch methods together while avoiding their disadvantages. We derive an iteratively Approximation-Shrink Scheme (AS Scheme) from the Maximum-A-Posterior (MAP) formulation using sequential approximation. We show that the state space can be effectively shrunk, but there may exist conflicts in the sequential optimization and the results may vary with different optimization sequences. In order to avoid these problems, an Ambiguity-Clearness Graph (A-C Graph) is formulated to efficiently represent the tracklet fragments and ambiguities in the association. A set of rules and procedures are defined for changes of nodes and edges in the graph, \eg, connections, disconnections, transforms, merges, \etc. A sliding Window-of-Ambiguity (WOA) is defined in the A-C Graph for sequential optimization of layers in the graph. Based on the A-C Graph and the sliding WOA optimization, MOT is conducted in a window-wise manner, which is able to disambiguate the association and accelerate the optimization process. We also show that the traditional online and batch approach can be embraced into this framework with different window sizes.

Our main contributions can be summarized as: (1) an approximation-shrink scheme that iteratively approximate the global optimization, (2) a window-wise optimization framework based on the novel A-C Graph which embrace the traditional online and batch methods, (3) a unified analysis of window-wise approaches with different window sizes using search tree.

\section{Related Works}
\label{sec:related}
Different from the past tracking methods~\cite{Mutihypothesistracking1979,JPDA1983}, TBD reconstructs trajectories of targets by associating detections provided by the object detectors. Most of the researchers exploits the TBD framework to design their algorithms in MOT, which can be categorized as online and batch approaches.

As for batch tracking~\cite{Milan2014,Milan2013,Milan2012,Butt2013,Segal2013,Dicle2013,MonteCarlo3Dtrackingstrategy2011} approaches, conditional random field (CRF) is often used to learn and model the affinity such as appearance and motion to discriminate among different trajectories~\cite{Yang2012,Yang2014}. A global and pairwise model is learned online in ~\cite{Yang2014} to form an energy function, which is minimized offline via heuristic search. Despite the popularity of CRF model, extensive training is needed. Continuous energy model is introduced by a series of work~\cite{Milan2014,Milan2013,Milan2012}. Milan \etal \cite{Milan2014} built a comprehensive continuous energy function by linearly combining terms regarding appearance, motion, mutual exclusion, trajectory persistence, \etc. The continuous energy functions are easier to optimize than discrete ones, whereas they possess too many parameters and are hard to be tuned. Network flow is first applied to tracking by Zhang \etal~\cite{Zhang2008}. A graph is formed with states of targets as nodes and the associations as edges. The likelihood of the states are represented as the capacity of edges. Butt \etal~\cite{Butt2013} improved the network structure by defining their node as a candidate pair of matching observations between consecutive frames. In order for a better model of occlusions, \cite{Segal2013} designed a latent data association framework. Instead of assigning each detection to a corresponding track, they assume each detection is its own track and assign a latent data to each node to represent the association. In addition to the general modeling of targets, some people worked on tracking targets with specific characteristics, \eg, Dicle \etal~\cite{Dicle2013} focus on tracking targets with similar appearance but different motion patterns.

Online tracking~\cite{Bae2014,Bae2014TIP,Shitrit2014,Breitenstein2011,FCimformationtheoretic2015,Motioncontext2015,sparserepresentation2013} has become more and more popular these days. Network flow has also been adopted in online tracking. \cite{Shitrit2014} formulate multi-object tracking into a multi-commodity network flow problem. They use sparse appearance to reduce computational complexity. Lu \etal~\cite{sparserepresentation2013} constructed a dictionary using already tracked objects and assigned the new detections by minimizing the L1 regularized function. Wang \etal~\cite{Leastabsoluted2014} finds that the representation residuals follow the Laplacian distribution, by which they improved the sparse representation method on tracking. Hungarian algorithm is firstly introduced into tracking problems by Joo \etal~\cite{Joo2007} to solve the bipartite graph model they proposed. The frame-by-frame scheme of online tracking takes great advantages of hungarian algorithm. Bae \etal~\cite{Bae2014} designed tracklet confidence by considering the length, occlusion and affinity. Different strategies are applied to tracklets with high and low confidence. Hungarian algorithm is employed in the association for local and global association respectively. Hungarian algorithm greedily associates detections in consecutive frames which could possibly misses the global optimal and cause identity switches.  Besides the popularity of Hungarian algorithm in association algorithms, Bayesian framework is also one of the most popular model for target modeling. Bae \etal~\cite{Bae2014TIP} improved their previous work~\cite{Bae2014} by perform data association with a track existence probability, the provided detections are associated to the existed tracks and the corresponding track existence probabilities will be updated. Yoon \etal~\cite{Motioncontext2015} constructed a Relative Motion Network(RMN) to factor out the camera motion by considering motion context from multiple object and incorporate relative motion network to Bayesian framework.

\section{Approximate-Shrink Scheme}
\label{sec:form}
Given observations $\mathbb{Z}_{1:t}=\{Z_{(\tau,i)}|\tau=1,2,\dots,t, i = 1,2,\dots,n_{\tau}\}$ of a real time video sequence, where $n_{\tau}$ denotes the number of observations in frame $\tau$, we assume: (1) each observation $Z_{(\tau,i)}$ corresponds to a state $X_{(\tau,i)}$ \cite{Segal2013}, (2) states in the same frame are independent, (3) some of the states are already clear given observations. The Maximum-a-Posterior (MAP) formulation of MOT is
\begin{equation}\label{eqn:MAP}
  \hat{\mathbb{X}}_{1:t}= \argmax_{\mathbb{X}_{1:t}}P(\mathbb{X}_{1:t}|\mathbb{Z}_{1:t}).
\end{equation}
Based on Assumption (2), we resolve $\mathbb{X}_{1:t}$ as
\begin{equation}\label{eqn:MAP1}
  \hat{\mathbb{X}}_{1:t}= \argmax_{\mathbb{X}_{1:t}}\prod_{\tau=1}^{t}\prod_{i=1}^{n_{\tau}}{P(X_{(\tau,i)}|\mathbb{Z}_{1:t},\mathbb{X}_{1:\tau-1})}.
\end{equation}
Assumption (3) offers us an intuition that there exist some states $\mathbb{X}^C_{1:t}=\{X_{(\tau',j)}|P(X_{(\tau',j)}|\mathbb{Z}_{1:t},\mathbb{X}_{1:\tau'-1}) \approx P(X_{(\tau',j)}|\mathbb{Z}_{1:t})\}$. Denote $\mathbb{X}^A_{1:t}= \mathbb{X}_{1:t}\setminus\mathbb{X}^C_{1:t}$. We name $\mathbb{X}^C_{1:t}$ Clear states (C states) and $\mathbb{X}^A_{1:t}$ Ambiguous states (A states). The global optimization in Equation \ref{eqn:MAP1} can be relaxed to
\begin{equation}\label{eqn:MAP2}
  \hat{\mathbb{X}}^A_{1:t} = \argmax_{\mathbb{X}^A_{1:t}}\prod_{\forall X_{(\tau,i)}\in  \mathbb{X}^A_{1:t} }{P(X_{(\tau,i)}|\mathbb{Z}_{1:t},\mathbb{X}_{1:\tau-1})},
\end{equation}
and
\begin{equation}\label{eqn:MAP3}
  \hat{X}_{(\tau',j)} = \argmax_{X_{(\tau',j)}}P(X_{(\tau',j)}|\mathbb{Z}_{1:t}),  \forall X_{(\tau',j)}\in\mathbb{X}^C_{1:t}.
\end{equation}
Doing these two optimization separately is an approximation to Equation \ref{eqn:MAP1}. First, we sequentially optimize every state $X_{(\tau',j)}$ in $\mathbb{X}^C_{1:t}$ (approximation step) via Equation \ref{eqn:MAP3}. Then we set $\mathbb{X}^C_{1:t}$ fixed as the evidence for $\mathbb{X}^A_{1:t}$, and derive Equation \ref{eqn:MAP2} to
\begin{multline}\label{eqn:MAP5}
  \hat{\mathbb{X}}^A_{1:t} = \argmax_{\mathbb{X}^A_{1:t}}\prod_{\forall X_{(\tau,i)}\in \mathbb{X}^A_{1:t} }{P(X_{(\tau,i)}|\mathbb{Z}_{1:t},\mathbb{X}^A_{1:\tau-1},\mathbb{X}^C_{1:t})}
\end{multline}
(shrink step). We iteratively find the $\mathbb{X}^{C'}_{1:t}\in\mathbb{X}^A_{1:t}$, $\mathbb{X}^{C'}_{1:t} = \{X_{(\tau',j)}|P(X_{(\tau',j)}|\mathbb{Z}_{1:t},\mathbb{X}^A_{1:\tau'-1},\mathbb{X}^C_{1:t}) \approx P(X_{(\tau',j)} | \mathbb{Z}_{1:t}, \mathbb{X}^C_{1:t}) \}$, let $\mathbb{X}^{C}_{1:t} = \mathbb{X}^{C'}_{1:t}$ and repeat the above steps to shrink the search space.

This Approximate-Shrink Scheme (A-S Scheme) iteratively search and narrow down the state space. $\mathbb{X}^C_{1:t}$ serve as nucleus of trajectories in the space which attract states to associate to them. Some nucleus merge together in the iteration to form longer tracklets during the iteration. However, the space is still too large, and the convergence is not guaranteed. More approximations are needed to accelerate the speed and ensure the convergence of this scheme. Moreover, it is necessary to design a data structure so as to avoid conflicts of associations of states in $\mathbb{X}^C_{1:t}$ and the effects of the sequence on the optimization results. Therefore, we propose a self-organizing A-C Graph and window-wise optimization framework to meet the demands in this regard.

\section{Window-wise Optimization for Tracking}
\label{sec:opt}
\begin{figure*}
\centering
\subfigure[Frame 11 to 17]{
\label{fig:flow1}\hspace{-0.12in}
\includegraphics[width=1.82in]{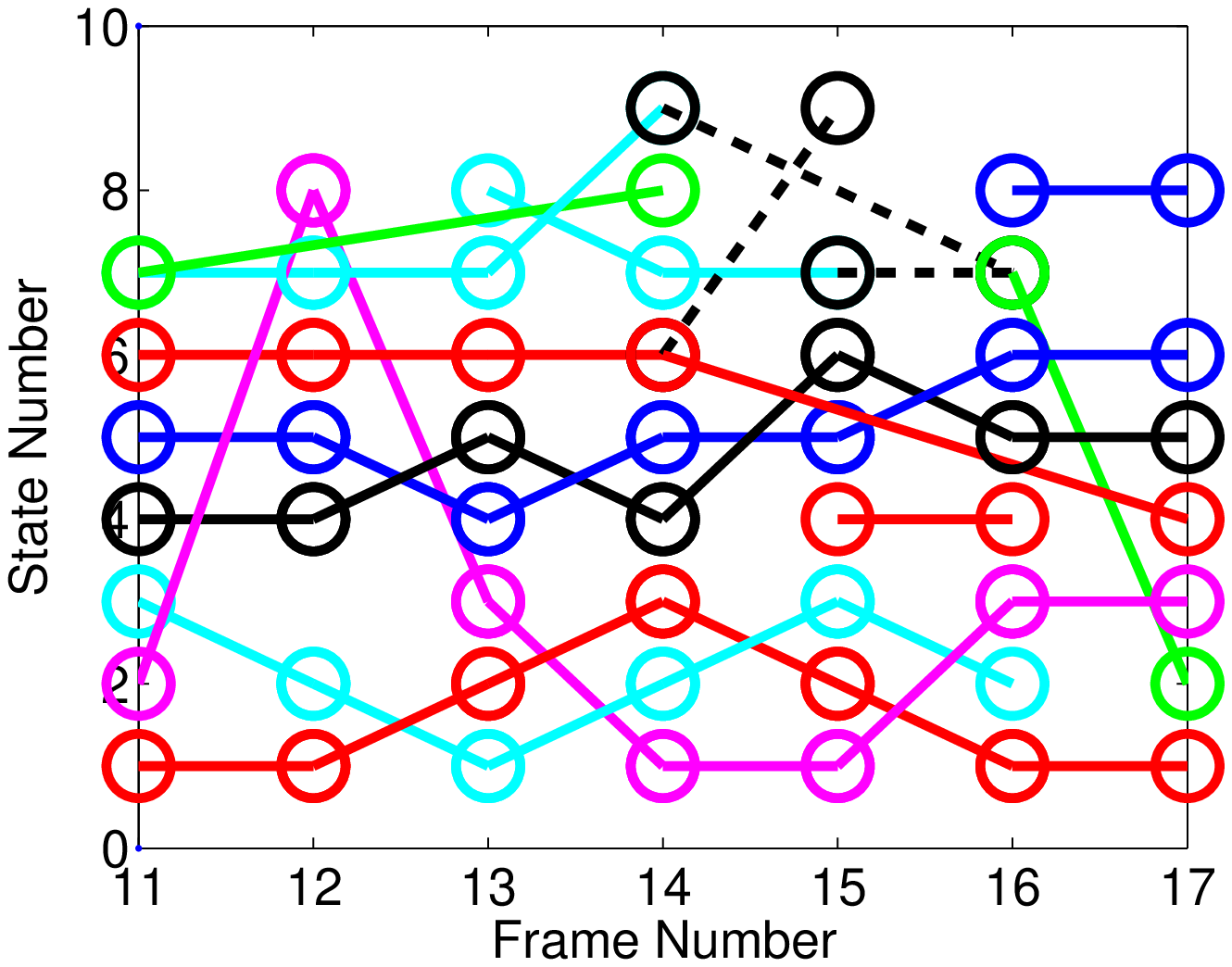}}\hspace{-0.19in}
\subfigure[Frame 12 to 18]{
\label{fig:flow2}
\includegraphics[width=1.82in]{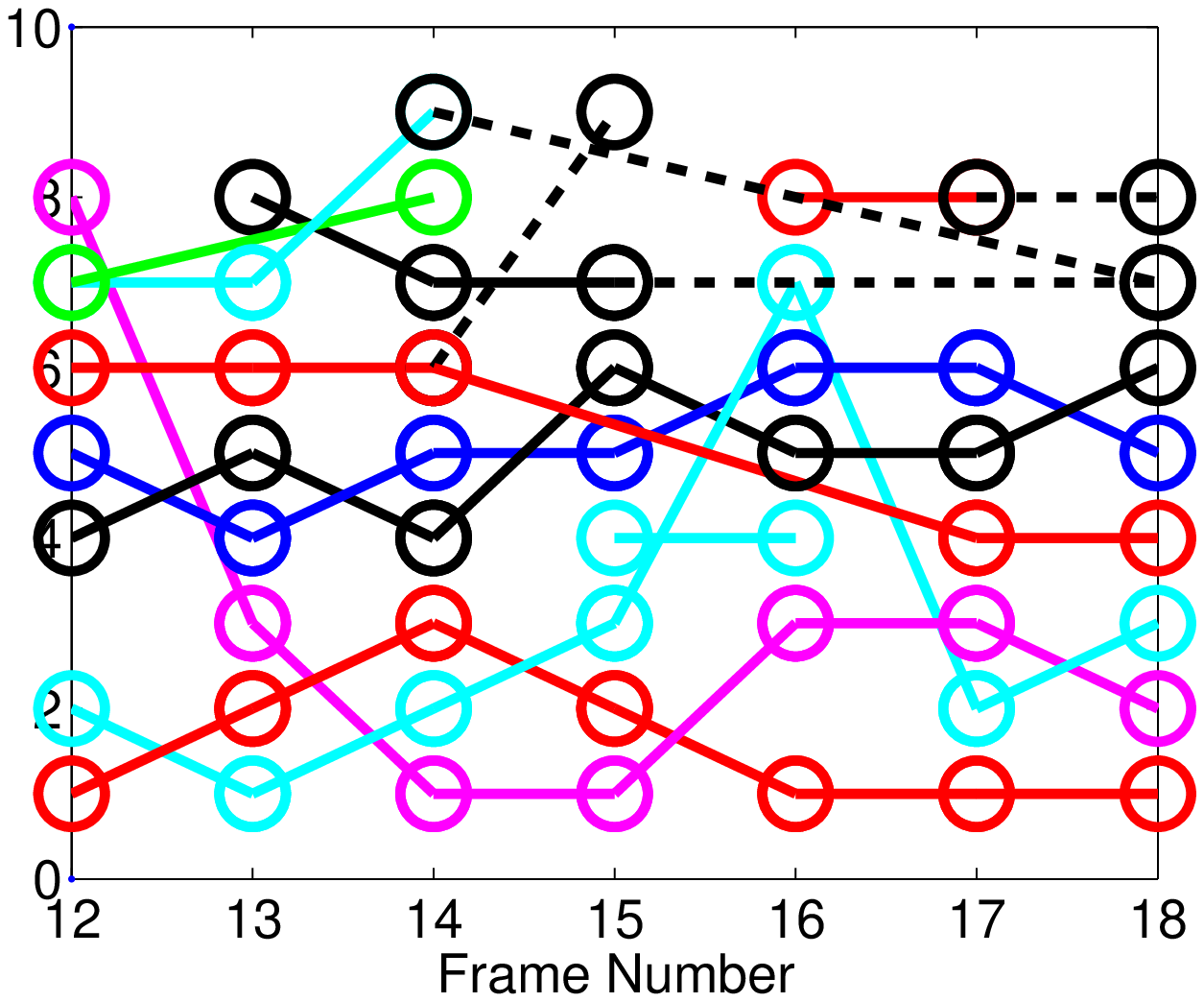}}\hspace{-0.19in}
\subfigure[Frame 13 to 19]{
\label{fig:flow3}
\includegraphics[width=1.82in]{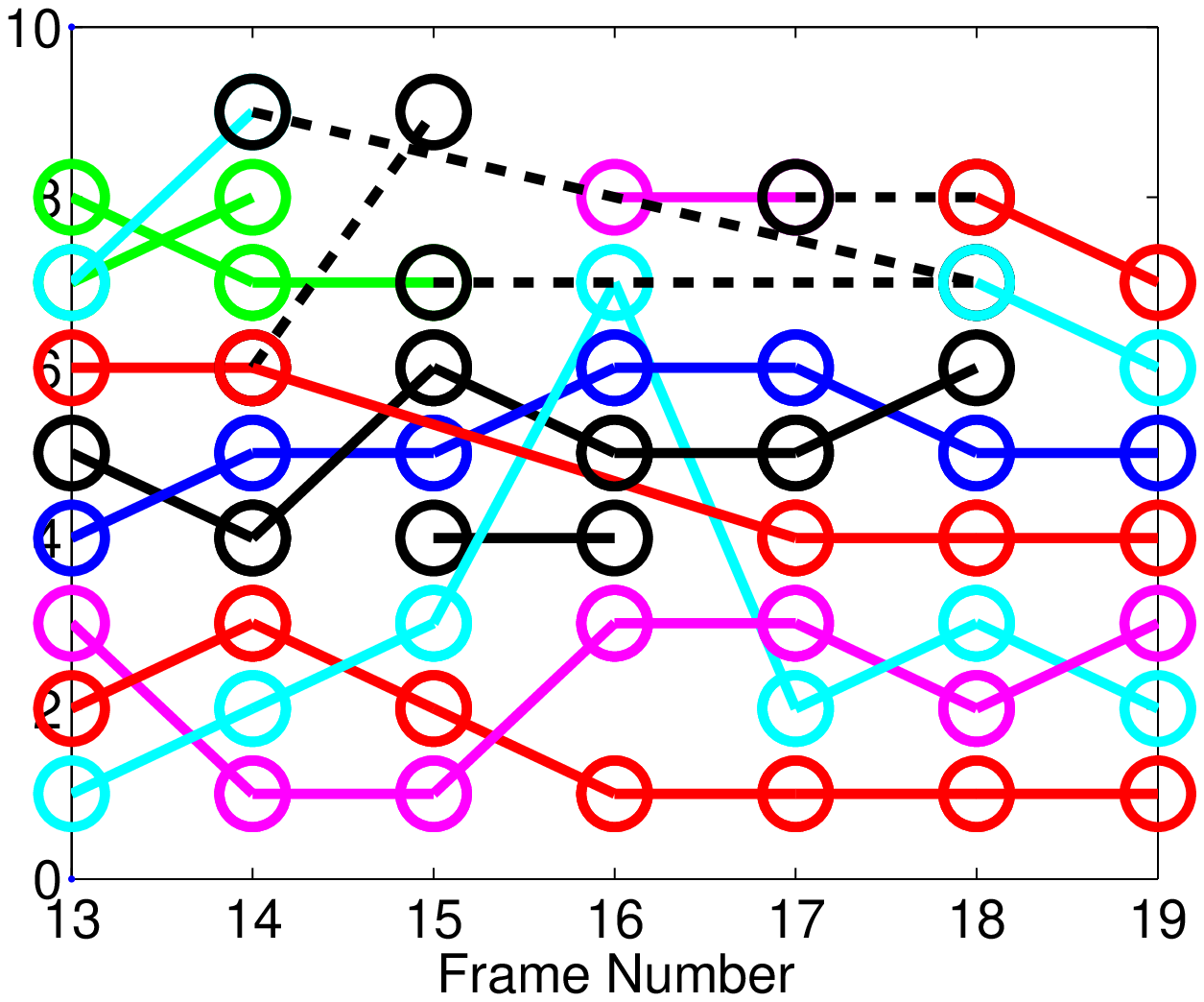}}\hspace{-0.19in}
\subfigure[Frame 14 to 20]{
\label{fig:flow4}
\includegraphics[width=1.82in]{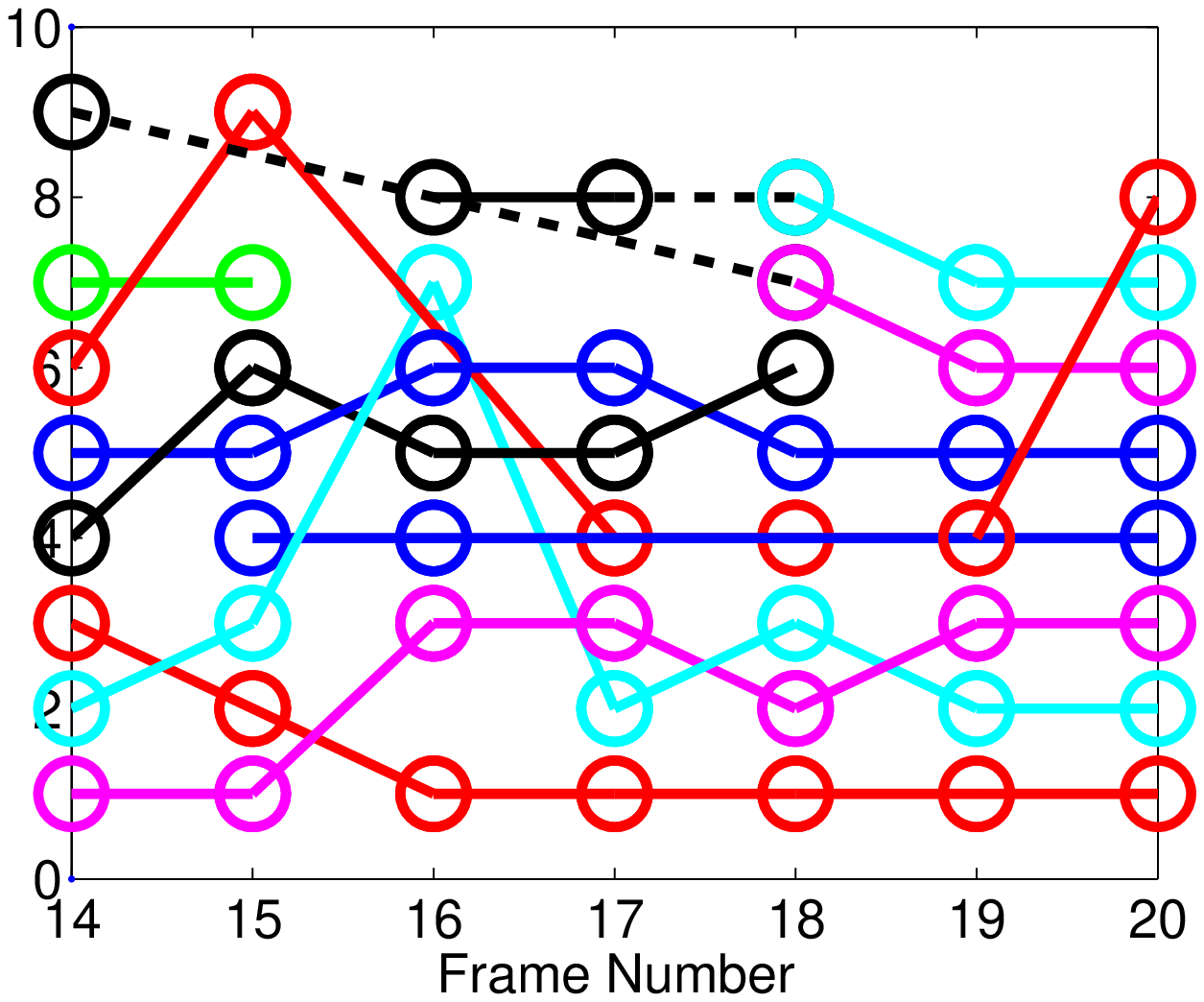}}
\label{fig:flow}
\caption{Visualization of the Window of Ambiguity (WOA) from frame 12 to frame 21 in TUD-Stadtmitte dataset. Each association is directed from parent to child and the A-C Graph is directed and acyclic. From (a) to (b), state $X_{(16,7)}$ was connected to state $X_{(14,2)}$ as a C state child, and was merged with $X_{(16,2)}$. Meanwhile, the association from $X_{(14,9)}$ and $X_{(14,7)}$ to $X_{(16,7)}$ were removed. From (c) to (d), state $X_{(15,9)}$ was inserted into the tracklet of state $X_{(14,6)}$ and state $X_{(17,4)}$. The figure is best shown in color.}
\end{figure*}

\subsection{Ambiguous-Clearness Graph}
{
\label{sec:def}
Given states $\mathbb{X}=\{X_{(\tau,i)}|\tau=1,2,\dots,t, i = 1,2,\dots,n_{\tau}\}$ and observations $\mathbb{Z}=\{Z_{(\tau,i)}|\tau=1,2,\dots,t, i = 1,2,\dots,n_{\tau}\}$ (the detections serve as observations in TBD multi-object tracking) in a real time video sequence, predefined thresholds $C_{\mathit{thre}}$ and $A_{\mathit{thre}}$ (the value of $C_{\mathit{thre}}$ and $A_{\mathit{thre}}$ are given in Section \ref{sec:exp}), we define state $X_{(\tau',j)}$ to be the \emph{parent} of state $X_{(\tau,i)}$ if $\tau'<\tau$ and there exists an association between $X_{(\tau',j)}$ and $X_{(\tau,i)}$, and $X_{(\tau,i)}$ is the \emph{child} of $X_{(\tau',j)}$. ($X_{(\tau,i)}$ and $X_{(\tau',j)}$ are only used as examples for clearness in illustration. They do not indicate certain states.) The \emph{determined parent} of a state is its only parent and the affinity score of the association is greater than $C_{\mathit{thre}}$. We now formally define the C states and A states. If a state $X_{(\tau,i)}$ has one determined parent or does not have parent, $X_{(\tau,i)}$ is a \emph{clear state} (C state), denoted as $X^{C}_{(\tau,i)}$. On the contrary, if $X_{(\tau,i)}$ has parent states but does not have a determined parent, it is an \emph{Ambiguous State} (A state), denoted as $X^{A}_{(\tau,i)}$. Note that a C state can only have zero or one parent. All the parents of a state $X_{(\tau,i)}$ form its \emph{active set}. We regulate that a state can have up to one C state as its child, and the frame number of its A state child should be smaller than that of its C state child. The observation corresponding to $X^{C}_{(\tau,i)}$ and $X^{A}_{(\tau,i)}$ is notated as $Z^{C}_{(\tau,i)}$ and $Z^{A}_{(\tau,i)}$. A \emph{clear association} is the association between a clear state and its parent, and a \emph{tracklet} is defined as a group states connected by clear association. The tracklet including $X_{(\tau,i)}$ is denoted as $\mathit{Trk}(X_{(\tau,i)})$. The C states in $\mathit{Trk}(X_{(\tau,i)})$ after $X_{(\tau,i)}$ is defined as the \emph{descendant} of $X_{(\tau,i)}$. By taking states and associations as the vertices and edges, we form the A-C Graph of the MOT problem. In this paper, we use states and associations instead of vertices and edges when discussing on the A-C Graph. The A-C Graph of TUD-Stadtmitte dataset is visualized in Figure \ref{fig:flow}, where the clear association is shown in solid line and the states belong to the same tracklet is in the same color.

As the association is directed from parent to child, the A-C Graph is a directed acyclic graph. In an A-C Graph, we define a time period $\tau'$ to $\tau$ ($1 \le \tau' < \tau \le t$) where there is only clear association in $1$ to $\tau'-1$ and $\tau+1$ to $t$ as Window-of-Ambiguity (WOA). The tracklet outside the WOA is determined and fixed and the changes of the states and association can only take place in the WOA. One can restrict the size of state space by setting the length of WOA.
}

\subsection{Actions}
{
As is mentioned in Section \ref{sec:form}, actions in A-C Graph should help avoid conflicts, \eg, multiple fathers for a C state, multiple C state children, clear association forms cycle, \etc. Meanwhile, the actions should be symmetrical to avoid the effect of chronological order. The basic actions of A-C Graph are initializations, disconnections, connections and merges between two states. Table \ref{tab:notation} shows functions and symbols used in defining these actions.

\begin{table}
  \footnotesize
  \centering
  \begin{tabular}{|c|c|}
  \hline
  Functions and Symbols & Description \\
  \hline
  \hline
  $\mathit{isempty}(\mathit{State Set})$ & Check whether the $\mathit{State Set}$ is empty. \\
  \hline
  $\mathit{find}(\mathit{State Set} = \mathit{C State})$ & Find the C States in the $\mathit{State Set}$.\\
  \hline
  $X_{(\tau,i)}.\mathit{pStates}$ & Find all the parents of $X_{(\tau,i)}$.\\
  \hline
  $X_{(\tau,i)}.\mathit{cldStates}$ & Find all the children of $X_{(\tau,i)}$.\\
  \hline
  $X_{(\tau,i)}.\mathit{frameN}$ & Find the frame number of $X_{(\tau,i)}$.\\
  \hline
  $X_{(\tau,i)}.\mathit{isClear}$ & Judge whether $X_{(\tau,i)}$ is a clear state.\\
  \hline
  \multirow{2}{*}{$X_{(\tau,i)}.\mathit{conf}$} & The affinity scores between all the\\
   & fathers of $X_{(\tau,i)}$ and $X_{(\tau,i)}$. \\
  \hline
  \end{tabular}
  \normalsize
  \caption{The functions and symbols used in this paper.}\label{tab:notation}
\end{table}

For a newly-entered state $X_{(\tau,i)}$, first we initialize the active set by enumerating all the potential parents. As is regulated in Section \ref{sec:def}, $X_{(\tau,i)}$ is able to connect with states in the previous frames, who does not have C state child or whose C state child is after $X_{(\tau,i)}$. Procedure \ref{alg:active} shows the pseudocode of initializing the active set.

We disconnect two states $X_{(\tau,i)}$ and $X_{(\tau',j)}$ by removing the association between them, and update these two states.

As is shown in Procedure \ref{alg:connectA}, we assign $X_{(\tau,i)}$ to $X_{(\tau',j)}$ as A state child. The procedure is terminated if $X_{(\tau,i)}$ is already a C state. If not, we check the descendant of $X_{(\tau',j)}$. If $X_{(\tau',j)}$ has no descendants, we directly add an association between $X_{(\tau,i)}$ and $X_{(\tau',j)}$, otherwise, we find the nearest C state descendant $X^p$ in the tracklet of $X_{(\tau',j)}$ not after $X_{(\tau,i)}$. If $X^p$ is in frame $\tau$, the procedure is terminated. If $X^p$ is before $X_{(\tau,i)}$, add the association between $X_{(\tau,i)}$ and $X^p$.

Procedure \ref{alg:connectC} illustrates the action that $X_{(\tau,i)}$ is connected to $X_{(\tau',j)}$ as C state child. If $X_{(\tau,i)}$ is currently not a C state, the existing parents of $X_{(\tau,i)}$ are removed. If $X_{(\tau',j)}$ does not have C state children, we directly add a connection between $X_{(\tau,i)}$ and $X_{(\tau',j)}$, otherwise, we find $X_{(\tau',j)}$'s latest C state descendant $X^p$ not after $X_{(\tau,i)}$. If $X^p$ is in frame $\tau$, $X^p$ and $X_{(\tau,i)}$ are merged together via Procedure \ref{alg:merge}. As $X^p$ is before $X_{(\tau,i)}$, an association is added between $X_{(\tau,i)}$ and $X^p$. All the A and C state children of $X^p$ after $X_{(\tau,i)}$ are removed from $X^p$ and reconnected to $X_{(\tau,i)}$ following Procedure \ref{alg:connectA} and \ref{alg:connectC} respectively. If $X_{(\tau,i)}$ is currently a C state and $X_{(\tau',j)}$ is not, $X_{(\tau',j)}$ is inserted into $X_{(\tau,i)}$'s tracklet using Procedure \ref{alg:connectC} if there is not a state in frame $\tau'$ in $X_{(\tau,i)}$'s tracklet and Procedure \ref{alg:merge} if there exists a state $X^{p1}$ in frame $\tau'$. If $X_{(\tau,i)}$ and $X_{(\tau',j)}$ are both C states, the two tracklets $X_{(\tau,i)}$ and $X_{(\tau',j)}$ will be grouped into one by recursively calling Procedure \ref{alg:connectC} and \ref{alg:merge}, as shown in Procedure \ref{alg:connectC}. If one of the two states is in a tracklet, the other state will be inserted into the tracklet.

Procedure \ref{alg:merge} describes the process of merging $X_{(\tau,j)}$ to $X_{(\tau,i)}$ in the same frame. As we cannot make changes on the states and tracklets outside WOA, we ensure that $X_{(\tau,j)}$ and $X_{(\tau,i)}$ cannot be C states at the same time to avoid merging of states outside WOA. For the descendants of $X_{(\tau,i)}$ and $X_{(\tau,j)}$, we recursively merge them into one tracklet by Procedure \ref{alg:connectC}. For the A state child $X^{\mathit{cld}}$ of $X_{(\tau,j)}$, we simply remove the association between $X^{\mathit{cld}}$ and $X_{(\tau,j)}$ and connect it to $X_{(\tau,i)}$ via Procedure \ref{alg:connectA}.

\begin{algorithm}\caption{Initialize the active set for the state $X_{(\tau,i)}$.}
\label{alg:active}
\begin{algorithmic}
\footnotesize
\Require state $X_{(\tau,i)}$, latest frame number $t$, size of Window of Ambiguity (WOA) $l$
\Ensure the active set containing all the potential parents of $X_{(\tau,i)}$
\For {all the states $X_{(\tau',j)}$ in frame $t-l+1$ to $\tau$}
\If {$\mathit{isempty}(\mathit{find}(X_{(\tau',j)}.\mathit{cldStates} = \mathit{CState}))$ \textbf{or} $\mathit{find}(X_{(\tau',j)}.\mathit{cldStates} = \mathit{CState}).frameN \le \tau$}
\State Add $X_{(\tau',j)}$ to the active set
\EndIf
\EndFor
\normalsize
\end{algorithmic}
\end{algorithm}

\begin{algorithm}\caption{Connect state $X_{(\tau,i)}$ to state $X_{(\tau',j)}$ as A state child.}
\label{alg:connectA}
\begin{algorithmic}
\footnotesize
\Require child state $X_{(\tau,i)}$, parent state $X_{(\tau',j)}$
\Ensure the updated network
\If {$X_{(\tau,i)}.\mathit{isClear}=\mathit{false}$}
\State $X^p=X_{(\tau',j)}$
\While {$($ \textbf{not} $\mathit{isempty}(\mathit{find}(X^p.\mathit{cldStates} = \mathit{CState}))$ \textbf{and} $X^p.\mathit{frameN}\le \tau$}
\State $X^p=$ the C state child of $X^p$
\EndWhile
\If {$X^p.\mathit{frameN}<\tau$}
\State Add $X_{(\tau,i)}$ to $X^p.\mathit{cldStates}$
\State Add $X^p$ to $X_{(\tau,i)}.\mathit{pStates}$
\State Update the features of $X_{(\tau,i)}$ and $X^p$
\EndIf
\EndIf
\normalsize
\end{algorithmic}
\end{algorithm}

\begin{algorithm}\caption{Connect state $X_{(\tau,i)}$ to state $X_{(\tau',j)}$ as C state child, where $X_{(\tau,i)}$ is currently an A state.}
\label{alg:connectC1}
\begin{algorithmic}
\footnotesize
\Require child state $X_{(\tau,i)}$, parent state $X_{(\tau',j)}$, latest frame number $t$, size of Window of Ambiguity (WOA) $l$
\Ensure the updated network
\State $X^p=X_{(\tau',j)}$
\While {$($ \textbf{not} $\mathit{isempty}(\mathit{find}(X^p.\mathit{cldStates} = \mathit{CState}))$ \textbf{and} $X^p.\mathit{frameN}\le \tau$}
\State $X^p=$ the C state child of $X^p$
\EndWhile
\If {$X^p.\mathit{frameN}=\tau$}
\State Do Procedure \ref{alg:merge} with $(X_{(\tau,i)}$,$X^p$,$t$,$l)$ as input
\Else
\State Remove all parents of $X_{(\tau,i)}$
\State Remove all children of $X^p$ in the same frame with $X_{(\tau,i)}$
\For {all children $X^{\mathit{cld}}$ of $X^p$ in the frames after $X_{(\tau,i)}$}
\State Remove the association between $X^{\mathit{cld}}$ and $X^p$
\If {$X^{\mathit{cld}}.\mathit{isClear}=\mathit{true}$}
\State Do Procedure \ref{alg:connectC} with $(X^{\mathit{cld}}$,$X_{(\tau,i)}$,$t$,$l)$ as input
\Else
\State Do Procedure \ref{alg:connectA} with $(X^{\mathit{cld}}$,$X_{(\tau,i)})$ as input
\EndIf
\EndFor
\EndIf
\normalsize
\end{algorithmic}
\end{algorithm}

\begin{algorithm}\caption{Connect state $X_{(\tau,i)}$ to state $X_{(\tau',j)}$ as C state child, where $X_{(\tau,i)}$ is currently a C state.}
\label{alg:connectC2}
\begin{algorithmic}
\footnotesize
\Require child state $X_{(\tau,i)}$, parent state $X_{(\tau',j)}$, latest frame number $t$, size of Window of Ambiguity (WOA) $l$
\Ensure the updated network
\State $X^{p1}=X_{(\tau',j)}$
\While {$(X^{p1}.\mathit{isClear} = \mathit{true})$ \textbf{and} $X^{p1}.\mathit{frameN} > t-l$}
\State $X^{p1}=$ the determined father of $X^{p1}$
\EndWhile
\State $X^{p2}=X_{(\tau',j)}$
\While {$(X^{p2}.\mathit{isClear} = \mathit{true})$ \textbf{and} $X^{p2}.\mathit{frameN} > t-l$}
\State $X^{p2}=$ the determined father of $X^{p2}$
\EndWhile
\If {$X^{p1}.\mathit{frameN} > X^{p2}.\mathit{frameN}$}
    \State Do Procedure \ref{alg:connectC} with $(X^{p2}$,$X^{p1}$,$t$,$l)$ as input
\ElsIf {$X^{p2}.\mathit{frameN} > X^{p1}.\mathit{frameN}$}
    \State Do Procedure \ref{alg:connectC} with $(X^{p1}$,$X^{p2}$,$t$,$l)$ as input
\Else
    \If {$X^{p1}.\mathit{isClear}=\mathit{true}$ \textbf{and} $X^{p2}.\mathit{isClear}=\mathit{true}$}
        \If {$X^{p1}.\mathit{conf}\ge X^{p2}.\mathit{conf}$}
        \State Remove the parent of $X^{p2}$, $X^{p2} = \mathit{A State}$
        \State Do Procedure \ref{alg:merge} with $(X^{p1}$,$X^{p2}$,$t$,$l)$ as input
        \Else
        \State Remove the parent of $X^{p1}$, $X^{p1} = \mathit{A State}$
        \State Do Procedure \ref{alg:merge} with $(X^{p2}$,$X^{p1}$,$t$,$l)$ as input
        \EndIf
    \ElsIf {$X^{p1}.\mathit{isClear}=\mathit{true}$}
        \State Do Procedure \ref{alg:merge} with $(X^{p1}$,$X^{p2}$,$t$,$l)$ as input
    \Else
        \State Do Procedure \ref{alg:merge} with $(X^{p2}$,$X^{p1}$,$t$,$l)$ as input
    \EndIf
\EndIf
\normalsize
\end{algorithmic}
\end{algorithm}

\begin{algorithm}\caption{Connect state $X_{(\tau,i)}$ to state $X_{(\tau',j)}$ as C state child.}
\label{alg:connectC}
\begin{algorithmic}
\footnotesize
\Require child state $X_{(\tau,i)}$, parent state $X_{(\tau',j)}$, latest frame number $t$, size of Window of Ambiguity (WOA) $l$
\Ensure the updated network
\If {$X_{(\tau,i)}.\mathit{isClear}=\mathit{false}$}
\State Do Procedure \ref{alg:connectC1} with $(X_{(\tau,i)}$,$X_{(\tau',j)}$,$t$,$l)$
\Else
\State Do Procedure \ref{alg:connectC2} with $(X_{(\tau,i)}$,$X_{(\tau',j)}$,$t$,$l)$
\EndIf
\normalsize
\end{algorithmic}
\end{algorithm}

\begin{algorithm}\caption{Merge state $X_{(\tau,j)}$ with state $X_{(\tau,i)}$.}
\label{alg:merge}
\begin{algorithmic}
\footnotesize
\Require state $X_{(\tau,i)}$, state $X_{(\tau,j)}$, latest frame number $t$, size of Window of Ambiguity (WOA) $l$
\Ensure the updated network
\If {$X_{(\tau,j)}.\mathit{isClear}$}
\State Remove $X_{(\tau,j)}$ from its parent $X^p$, if any
\State Do Procedure \ref{alg:connectC} with $(X_{(\tau,i)}$,$X^p$,$t$,$l)$ as input
\Else
\State Remove $X_{\tau}$ from its parents $X^p$
\State For all $X^p$, do Procedure \ref{alg:connectA} with $(X_{(\tau,i)}$,$X^p)$ as input
\EndIf
\For {all the children $X^{\mathit{cld}}$ of $X_{(\tau,j)}$}
\If {$X^{\mathit{cld}}.\mathit{isClear}=true$}
\State Remove $X^{\mathit{cld}}$ from $X_{(\tau,j)}$
\State Do Procedure \ref{alg:connectC} with $(X^{\mathit{cld}}$,$X_{(\tau,i)}$,$t$,$l)$ as input
\Else
\State Remove $X^{\mathit{cld}}$ from $X_{(\tau,j)}$
\State Do Procedure \ref{alg:connectA} with $(X^{\mathit{cld}}$,$X_{(\tau,i)}$,$t$,$l)$ as input
\EndIf
\EndFor
\normalsize
\end{algorithmic}
\end{algorithm}

Although there exists recursion in the actions, it can be easily proved that the recursion in Procedure \ref{alg:connectA}, \ref{alg:connectC} and \ref{alg:merge} cannot form an endless recursion loop, and the sequence of carrying out actions on a set of states will not affect the structure of A-C Graph. Visualization of these actions in TUD-Stadtmitte dataset can be found in Figure \ref{fig:flow}. In Figure \ref{fig:flow2}, newly-entered states $X_{(18,1)}$ to $X_{(18,8)}$ connected to their initial active sets via Procedure \ref{alg:active}, \ref{alg:connectA} and \ref{alg:connectC}. From Figure \ref{fig:flow1} to \ref{fig:flow2}, $X_{(16,7)}$ was connected to $X_{(14,2)}$ as a C state child by Procedure \ref{alg:connectC}, and merged with $X_{(16,2)}$ using Procedure \ref{alg:merge}.
}

\subsection{Sliding Window Optimization}
{
\label{sec:swo}
For a real time sequence, the A-C Graph is continuously adding new states from latest frame $t$. The WOA should be sliding to keep its size from being too large and remove the ambiguities to generate tracks. So we set the upper bound of the size of WOA as $l$.

The sliding window optimization consists of three steps. First, for all the newly-entered states $X_{(t,i)}$ in frame $t$, $i = 1,\dots,n_t$, we find the active sets via Procedure \ref{alg:active} and compute the affinity score $a(X_{(t,i)},X^p)$ between $X_{(t,i)}$ and each state $X^p$ in the corresponding active set. If $a(X_{(t,i)},X^p) \ge C_{thre}$, do Procedure \ref{alg:connectC} with $(X_{(t,i)},X^p,t,l)$ as input. If $A_{thre} < a(X_{(t,i)},X^p) < C_{thre}$, do Procedure \ref{alg:connectA} with $(X_{(t,i)},X^p)$ as input. Second, from frame $t-l$ to $t$, we sequentially recompute the affinity score of states in the same frame with their fathers and reconnect them according to the new affinity. Third, Hungarian Algorithm \cite{ahuja1988network} is carried out on states in frame $t-l$ with their father states to get the best arrangement of association and clear all the ambiguity in frame $t-l$. All states in frame $t-l$ are transformed to C states and the WOA shifts forward. If $t$ has not reached the end, $t = t + 1$ and return to the first step, otherwise, $l = l - 1$ and redo the third step. The outline of the optimization process is shown in Procedure \ref{alg:sliding}.

\begin{algorithm}\caption{Conduct sliding window optimization for MOT.}
\label{alg:sliding}
\begin{algorithmic}
\footnotesize
\Require size of Window-of-Ambiguity (WOA) $l$
\Ensure the final A-C Graph and the association result
\State 1. Associate the newly-entered states in the latest frame $t$ to their initial active sets.
\State 2. Sequentially shrink the active set of each A state in WOA.
\State 3. Determine the association of states in frame $t-l$ using Hungarian Algorithm \cite{ahuja1988network}.
\If {$t$ has not reached the end}
\State $t=t+1$, return to 1
\Else
\State $l=l+1$, return to 3
\EndIf
\normalsize
\end{algorithmic}
\end{algorithm}

The sliding window optimization conducts A-S Scheme in a window-wise manner. Procedure \ref{alg:connectC} and \ref{alg:merge} in step one and two serve as the approximation step, and updating affinity score in step two follows the shrink step. Step three forces the states in frame $t-l$ to determine their connections, which guarantees the convergence.
}

\subsection{Online, Delayed and Batch Methods}
\begin{figure*}
\centering
\subfigure[Search Space]{
\label{fig:tree0}
\includegraphics[width=1.49in]{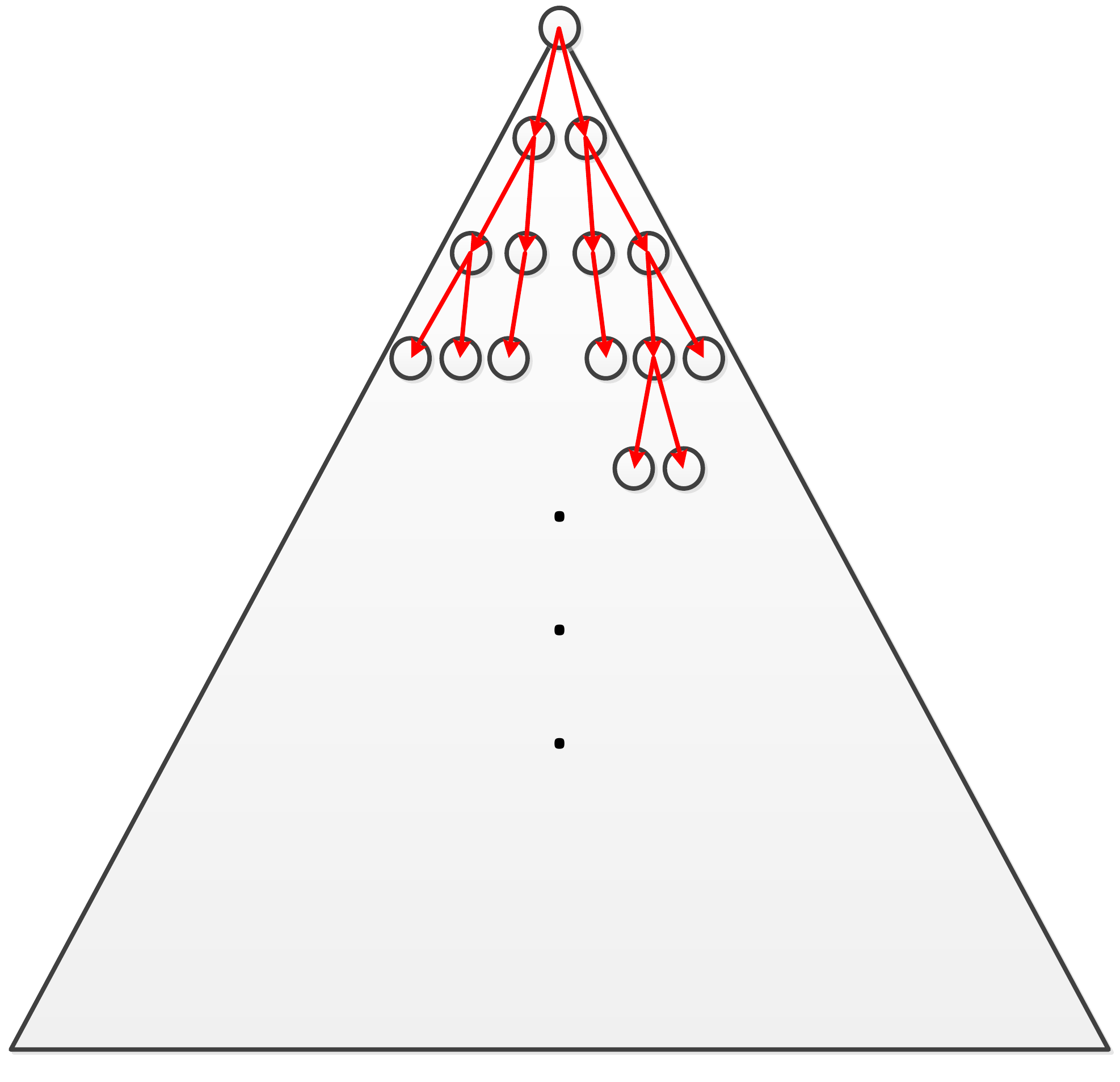}}
\subfigure[Online $l=1$]{
\label{fig:tree1}
\includegraphics[width=1.53in]{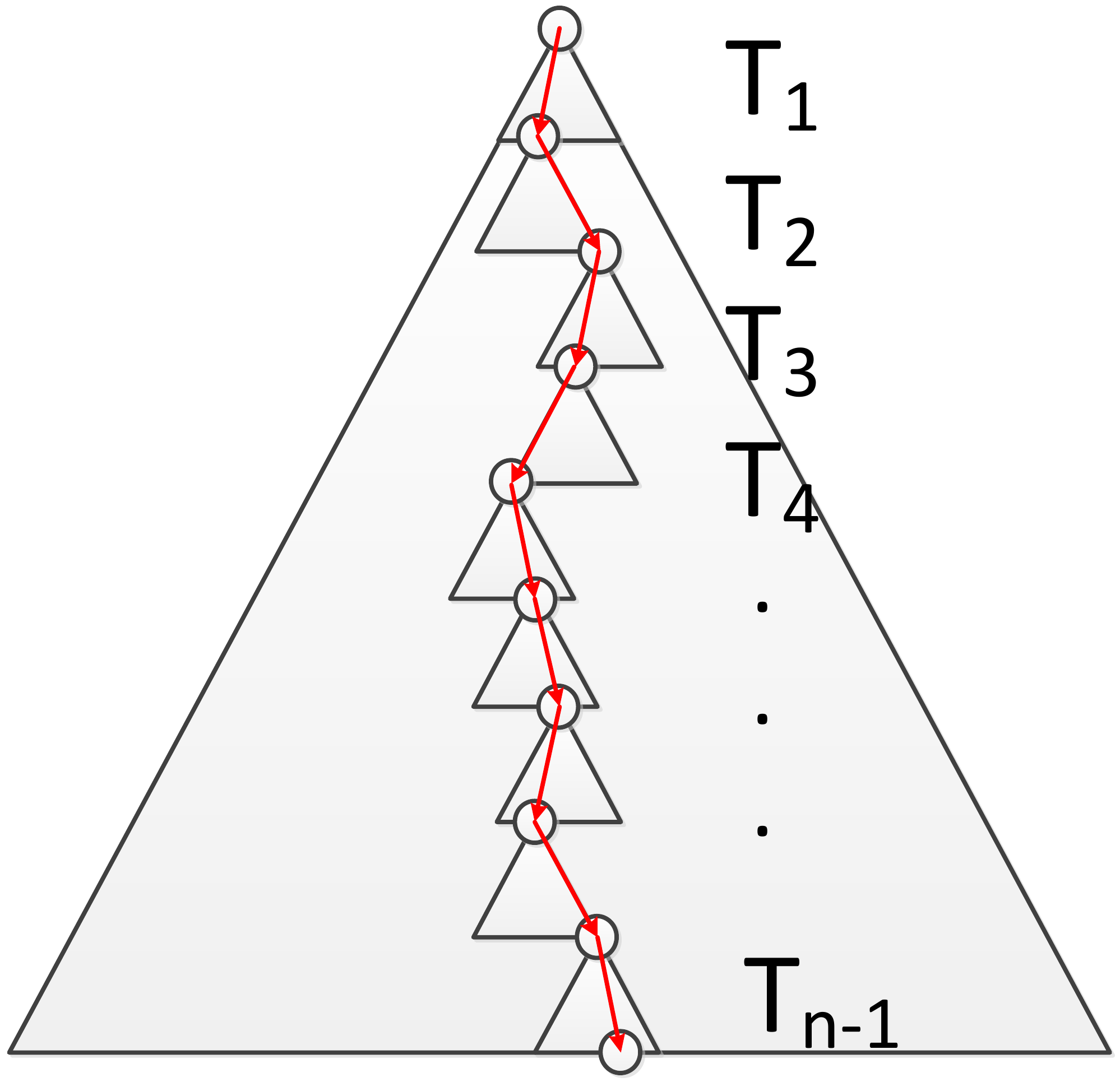}}
\subfigure[Delayed $1<l<t$]{
\label{fig:tree2}
\includegraphics[width=1.5in]{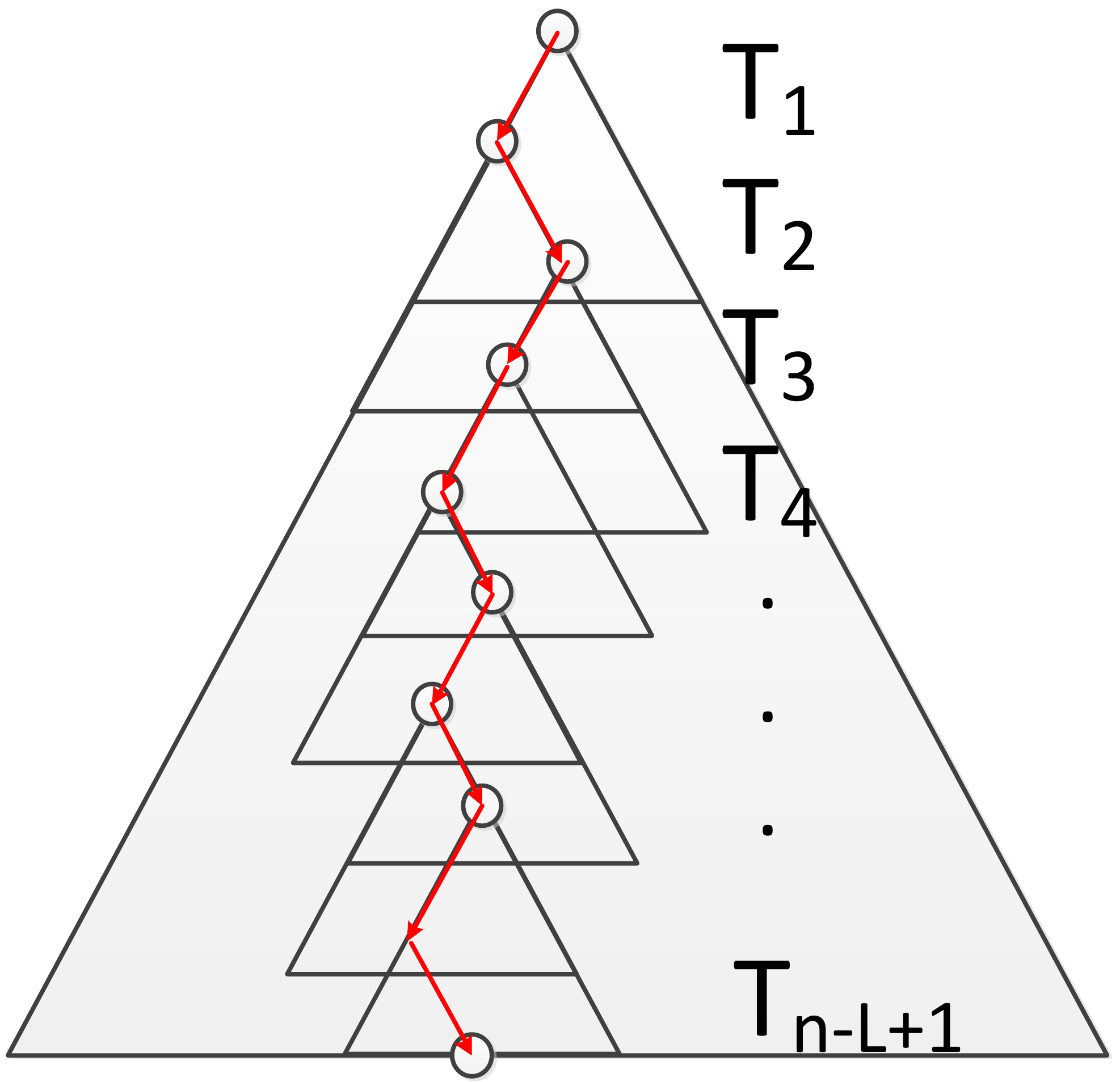}}
\subfigure[Batch $l=t$]{
\label{fig:tree3}
\includegraphics[width=1.5in]{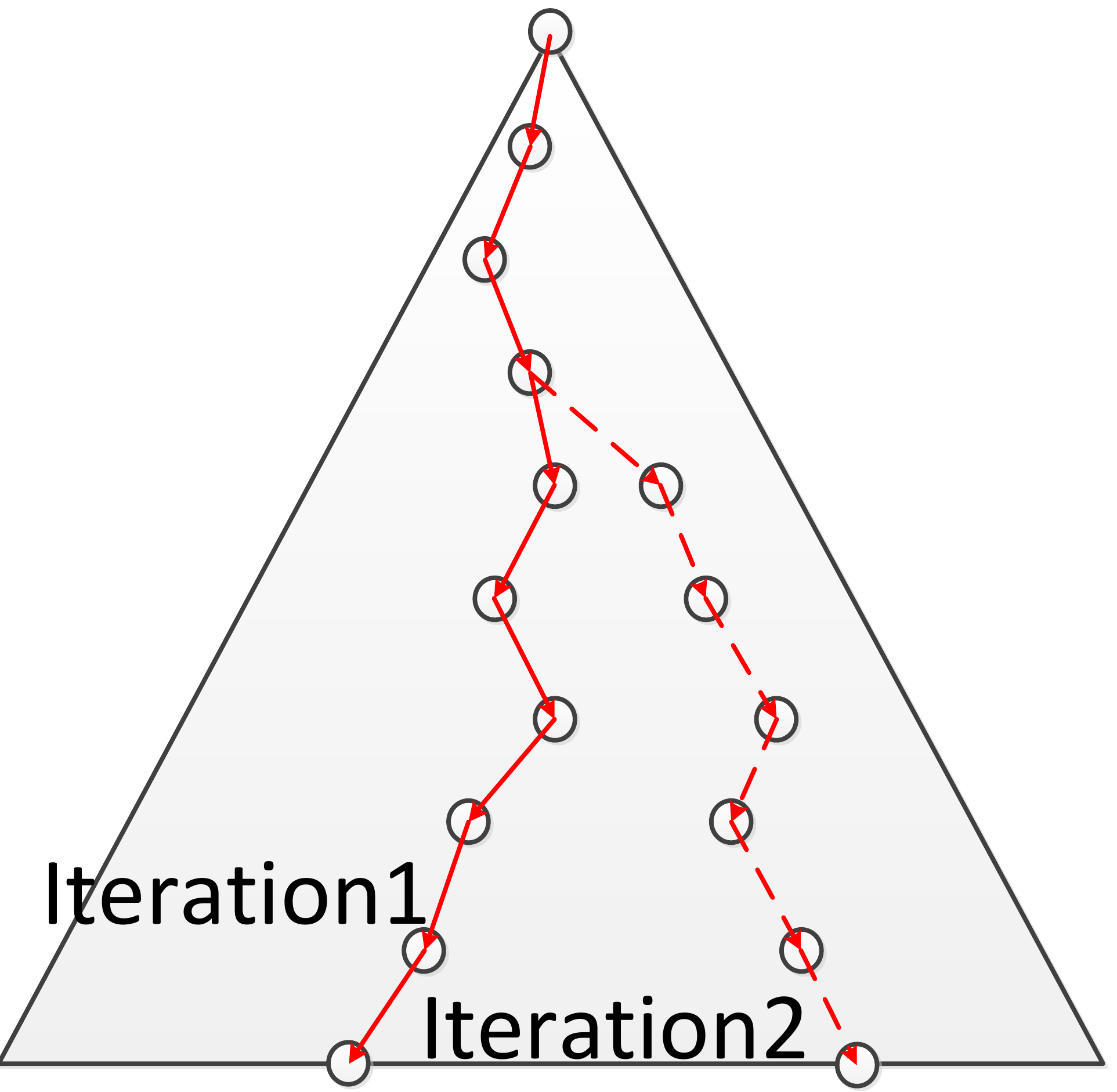}}
\caption{Formation of a trajectory with different $l$. (a) illustrates the original search space. (b),(c) and (d) stand for the search process with local search tree. $T_{\tau}$ indicates the SWO in frame $\tau$, $\tau = 1,\dots,(t-l+1)$. The red lines represent the associations. For online and delayed approaches, the trajectory are formed from top to bottom, while for batch approaches, the trajectory is formed and optimized via iteration.}
\label{fig:tree}
\end{figure*}

{
\label{sec:ODB}
Based on the definition of A-C Graph and sliding window optimization, we form this window-wise framework which includes online ($l = 1$), delay ($1<l<t$) and batch methods ($l = t$). Figure \ref{fig:tree} demonstrates the formation of a trajectory starting from $X^s$ in the A-C Graph via these three methods. The window-wise optimization finds a relatively small search tree $T_1,\dots,T_{t-l+1}$ according to $l$ at each iteration. As for an online method (Figure \ref{fig:tree1}), $l = 1$ and the search is greedy. For a delayed method (Figure \ref{fig:tree2}), heuristic search is conducted in $T_1,\dots,T_{t-l+1}$. The search space remains unchanged for a batch method (Figure \ref{fig:tree3}), so local search methods, \eg, hill climbing, simulated annealing, \etc, is often exploited to direct to local optimal iteratively. The experimental analysis of the relation between $l$ and optimization results is provided in Section \ref{sec:AWOA}.

}

\section{Experimental Evaluation}
\label{sec:exp}
\subsection{Implementation}
{
\label{sec:implem}
\textbf{Affinity model}: We implemented a basic affinity model, following \cite{Bae2014}, which includes the appearance model $\mathit{App}(X_{(\tau,i)},X_{(\tau',j)})$, motion model $\mathit{Mot}(X_{(\tau,i)},X_{(\tau',j)})$ and shape model $\mathit{Shp}(X_{(\tau,i)},X_{(\tau',j)})$. The appearance model measures the Bhattacharyya distance of histograms of $X_{(\tau,i)}$ and $X_{(\tau',j)}$. If $X_{(\tau,i)}$ is in a tracklet $\mathit{Trk}(X_{(\tau,i)})$, instead of using Incremental Linear Discriminant Analysis (ILDA) used in \cite{Bae2014}, we simply average the appearance histograms of all states in $\mathit{Trk}(X_{(\tau,i)})$ using an exponential discount factor. First-order Kalman filter is applied to smoothing and predicting positions of the targets and shapes of the bounding boxes. We compute the normalized distance of target positions and bounding box shapes and map them to a Gaussian distribution $N(O,\mathit{Var})$ to get the affinity scores. The overall affinity
\begin{equation}\label{eqn:aff1}
  \mathit{Aff}(X_{(\tau,i)},X_{(\tau',j)}) = \mathit{App}\times\mathit{Mot}\times\mathit{Shp}.
\end{equation}

\textbf{Dataset description}: We use the MOT Benchmark \cite{leal2015motchallenge} for training and evaluation in this paper, where the benchmark contains both $11$ sequences for training and testing. In total, there are $11,286$ frames, $5,503$ for training set and $5,783$ for testing set. The sequences possess different frame rates and resolutions, and only tracking pedestrians.

\textbf{Parameter Settings}: In our experiment, the $C_{thre} = 0.5$ and $A_{thre} = 0.1$. We estimate the length of every occlusion (number of frames with overlap $>$ 0.4) in the training set of MOT Benchmark and study the distribution of occlusion lengths. As shown in Figure \ref{fig:overlap},  about $99\%$ of the overlaps are within $5s$, and $84\%$ of which are within $1s$. Therefore, the delayed time is set to $1s$ and the length of WOA $l = $ frame rate $\times$ delayed time. The variance of the Gaussian distribution in the motion model and shape model is $ \mathit{Var} = [20^2, 50^2]$. Other parameters of the affinity model are the same as \cite{Bae2014}.

\begin{figure}
  \centering
  \includegraphics[width=3.2in]{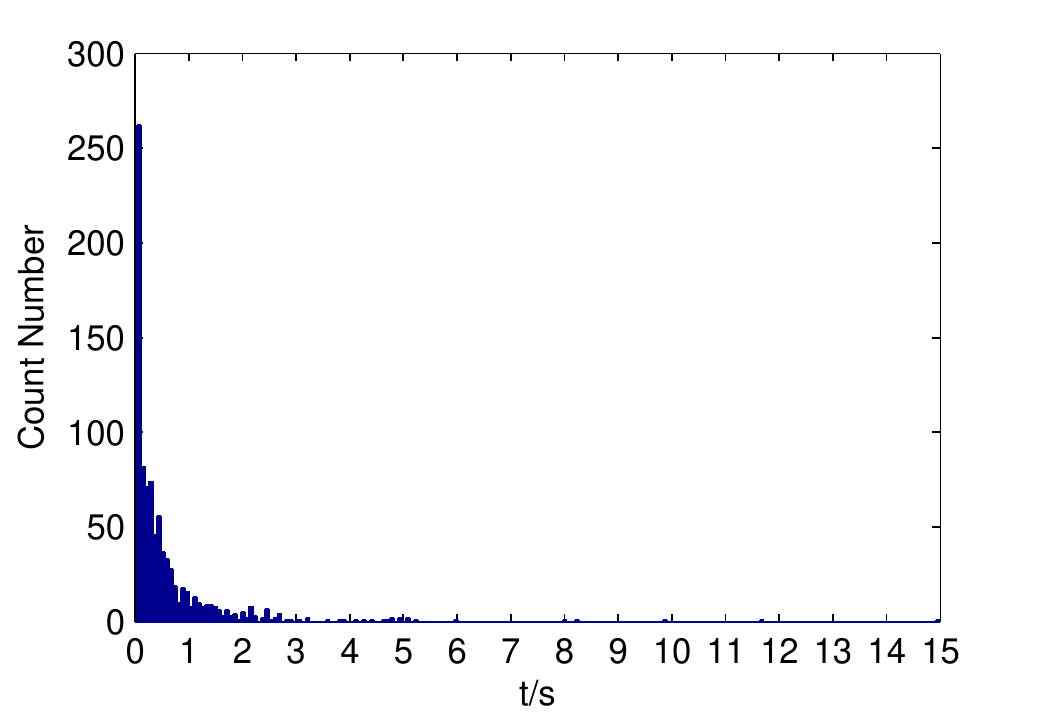}\\
  \caption{Distribution of lengths of bounding box overlaps in the ground truth sequences in MOT Benchmark \cite{leal2015motchallenge}. $99\%$ of the overlaps are within 5s and $84\%$ of them are in 1s.}\label{fig:overlap}
\end{figure}
}
\subsection{Analysis of Window-of-Ambiguity}
{
\label{sec:AWOA}
To analyze the connection of WOA size and the quality of the window-wise optimization, we define the energy of an A-C Graph as
\begin{equation}\label{eqn:energy}
  E(t) = -\sum_{\tau=1}^{t}\sum_{i=1}^{n_{\tau}}\max_{X^p \in X_{(\tau,i)}.pStates}\mathit{Aff(X_{\tau,i},X^p)}.
\end{equation}
Figure \ref{fig:energy} presents the final energy with varying size of WOA on TUD-Stadtmitte (number of frame $=179$), TUD-Campus (number of frame $=71$) and PETS-S2L2 (number of frame $=436$) in MOT Benchmark. The X-axis is in logarithmic scale. Interestingly, final energy of these sequences reduced significantly when window size $l$ grows from $1$ to $5$, while the speed of decrease become much slower when $l > 5$. Settings of these sequences, \eg, target density, viewpoint, \etc, are different, but the patterns of energy change almost remain identical. It is likely that the trend of final energy only deals with WOA size $l$. And the tracking results can be much improved with a small WOA comparing to the online method, which experimentally illustrates the better performance of delayed methods than online ones in the window-wise optimization framework. The final energy does not reduce too much when $l$ grows larger than $5$. This indicates the sliding window approximation only has a minor effect on the final performance. And it becomes a trade-off between speed and better results when WOA grows larger.
\begin{figure}
\centering
\subfigure[TUD-Stadtmitte(number of frame $=179$)]{
\label{fig:energy1}
\includegraphics[width=2.7in]{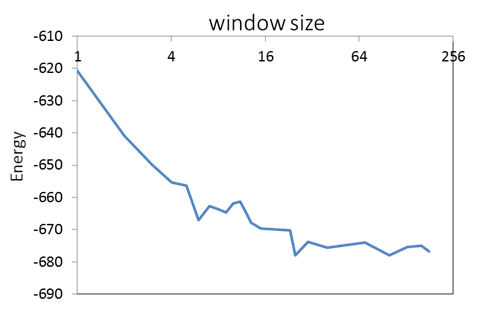}}
\subfigure[TUD-Campus(number of frame $=71$)]{
\label{fig:energy2}
\includegraphics[width=2.7in]{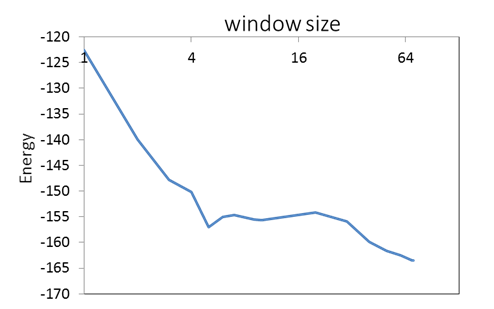}}
\subfigure[PETS-S2L2(number of frame $=436$)]{
\label{fig:energy3}
\includegraphics[width=2.7in]{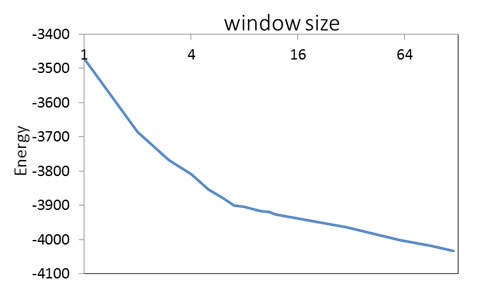}}
\caption{The final energy with varying size of Window-of-Ambiguity (WOA) on different sequences. The X-axis is in logarithmic-scale. The energy decrease rapidly when $l$ grows from $1$ to $5$. When $l>5$, the decrease of energy becomes slower.}
\label{fig:energy}
\end{figure}
}
\subsection{Performance Evaluation}
{
\label{sec:per}
\textbf{Evaluation Metrics:} We apply the CLEAR MOT \cite{keni2008evaluating} and \cite{Yang2012,kuopersonidentity2011}'s metric when evaluating our result. The multiple object tracking accuracy (MOTA) shows the combined accuracy based on the number of false positives (FP), identity switches (IDS) and missed targets (FN). The multiple object tracking precision (MOTP) measures the overlap of bounding boxes between ground truths and results given by trackers. MT and ML indicate the number of mostly tracked and lost targets. FG represents the number of fragmented tracks.

\textbf{Evaluation:} As shown in Table \ref{tab:result}, our method clearly outperforms the TC\_ODAL method using the same affinity model, not only in MOTA. Even in some datasets, shown in Table \ref{tab:dataset}, our method with the basic affinity model reached the performance of the methods using state-of-the-art affinity models.

\begin{table*}
\centering
\footnotesize
\label{tab:result}
\begin{tabular}{|c|c|c|c|c|c|c|c|c|c|}
  \hline
  Method & Type & MOTA$\uparrow$ & MOTP$\uparrow$ & MT$\uparrow$ & ML$\downarrow$ & FP$\downarrow$ & FN$\downarrow$ & IDS$\downarrow$ & FG$\downarrow$ \\
  \hline
  \hline
  AC-MOT(Proposed $+$ affinity of \cite{Bae2014}$\dagger$) & Delayed & $18.1\pm17.9$ & $70.4$ & $5.8\%$ &  $58.3\%$ &  $13,492$ &  $36,295$ &  $509$ & $1,092$  \\
  \hline
  TBD\cite{geiger20143d} & Batch & $15.9\pm17.6$ & $70.9$ & $6.4\%$ &  $47.9\%$ &  $14,943$ &  $34,777$ &  $1,939$ & $1,963$  \\
  \hline
  TC\_ODAL \cite{Bae2014}$\dagger$ & Online & $15.1\pm15.0$ & $70.5$ & $3.2\%$ &  $55.8\%$ &  $12,970$ &  $38,538$ &  $637$ & $1,716$  \\
  \hline
  DP\_NMS \cite{pirsiavash2011globally} & Batch & $14.5\pm13.9$ & $70.8$ & $2.3\%$ &  $40.8\%$ &  $13,171$ &  $34,814$ &  $4,537$ & $3,090$  \\
  \hline
\end{tabular}
\normalsize
\caption{Performance evaluation. Results can be found in \url{http://motchallenge.net/results/2D_MOT_2015/}. The best outcomes are marked in bold. $\uparrow$ represents higher is better, while $\downarrow$ stands for lower being better. Methods evaluated using the same set of affinity descriptor are marked with the same symbol.}
\end{table*}

\begin{table}
\centering
\footnotesize
\label{tab:dataset}
\begin{tabular}{|c|c|c|c|c|}
  \hline
  \multirow{2}*{Method} & \multirow{2}*{AC-MOT} & CEM & MotiCon & SegTrack\ \\
   & & \cite{Milan2014} & \cite{leal2014learning} & \cite{milanjoint} \\
  \hline
  type & Delayed & Batch & Batch & Batch \\
  \hline
  \hline
  TUD-Crossing & $62.3$ & $61.6$ & $58.2$ & $53.9$  \\
  \hline
  ETH-Linthescher & $18.2$ & $18.4$ & $18.3$ & $11.1$ \\
  \hline
  ETH-Crossing & $23.4$ & $18.2$ & $22.8$ & $23.4$ \\
  \hline
  KITTI-16 & $38.1$ & $31.6$ & 38.8 & 40.2 \\
  \hline
\end{tabular}
\normalsize
\caption{MOTA of some sequences in MOT Benchmark. We compare our method using \cite{Bae2014}'s affinity model with state-of-the-art affinity models.}
\end{table}
}

\section{Conclusion}
\label{sec:con}
This paper proposed an A-S Scheme for sequential approximation and a window-wise optimization framework based on the A-C Graph. The core idea of this method is to cluster the states subject to several constraints, \eg states in the same frame cannot be clustered into one group, \etc. The A-C Graph together with the sliding window optimization transformed the global clustering into a sequential local clustering which self-organized the structure in a relatively small state space, which can be done efficiently with little harm to handling occlusions. We showed experimentally that the characteristics of window-wise optimization framework rarely change with the varying settings of the sequence. As the affinity model serves as the distance metric in clustering, it can influence the results of clustering. Therefore, it is a fair comparison of optimization models if similar affinity models are adopted. The experimental results show that by using the basic affinity model, our method even showed competitive performance in an unfair test. Our future work is to realize more state-of-the-art affinity models to the window-wise optimization model. Also, we plan to design a unity interface, which can help to embed the affinity models into different optimization models much easier than now.

{\small
\bibliographystyle{ieee}
\bibliography{mot}
}

\end{document}